\documentclass[sigconf]{aamas} 
\renewcommand\footnotetextcopyrightpermission[1]{} 
\usepackage[noend]{algpseudocode}
\usepackage{algorithmicx,algorithm}
\usepackage{subfigure}
\usepackage{array}
\newtheorem{lem}{Lemma}

\usepackage{balance} 

\setcopyright{ifaamas}
\acmConference[AAMAS '23]{Proc.\@ of the 22nd International Conference
on Autonomous Agents and Multiagent Systems (AAMAS 2023)}{May 29 -- June 2, 2023}
{London, United Kingdom}{A.~Ricci, W.~Yeoh, N.~Agmon, B.~An (eds.)}
\copyrightyear{2023}
\acmYear{2023}
\acmDOI{}
\acmPrice{}
\acmISBN{}

\acmSubmissionID{243}

\title[AAMAS-2023 Formatting Instructions]{Centralized Cooperative Exploration Policy for Continuous Control Tasks}

\author{Chao Li*}
\affiliation{Institute of Automation, Chinese Academy of Sciences
\country{China}}
\email{lichao2021@ia.ac.cn}

\author{Chen Gong}
\affiliation{Institute of Automation, Chinese Academy of Sciences
\country{China}}
\email{gongchen2020@ia.ac.cn}
\authornote{These two authors contributed equally.}

\author{Qiang He}
\affiliation{University of Tubingen
\country{Germany}}
\email{qianghe97@gmail.com}

\author{Xinwen Hou}
\affiliation{Institute of Automation, Chinese Academy of Sciences
\city{Beijing}
\country{China}}
\email{xinwen.hou@ia.ac.cn}
\authornote{The corresponding author.}

\author{Yu Liu}
\affiliation{Institute of Automation, Chinese Academy of Sciences
\city{Beijing}
\country{China}}
\email{yu.liu@ia.ac.cn}

\begin{abstract}

The deep reinforcement learning (DRL) algorithm works brilliantly on solving various complex control tasks. This phenomenal success can be partly attributed to DRL encouraging intelligent agents to sufficiently explore the environment and collect diverse experiences during the agent training process. Therefore, exploration plays a significant role in accessing an optimal policy for DRL. Despite recent works making great progress in continuous control tasks, exploration in these tasks has remained insufficiently investigated. To explicitly encourage exploration in continuous control tasks, we propose CCEP (\textbf{C}entralized \textbf{C}ooperative \textbf{E}xploration \textbf{P}olicy), which utilizes underestimation and overestimation of value functions to maintain the capacity of exploration. CCEP first keeps two value functions initialized with different parameters, and generates diverse policies with multiple exploration styles from a pair of value functions. In addition, a centralized policy framework ensures that CCEP achieves message delivery between multiple policies, furthermore contributing to exploring the environment cooperatively. Extensive experimental results demonstrate that CCEP achieves higher exploration capacity. Empirical analysis shows diverse exploration styles in the learned policies by CCEP, reaping benefits in more exploration regions. And this exploration capacity of CCEP ensures it outperforms the current state-of-the-art methods across multiple continuous control tasks shown in experiments. 

\end{abstract}
\keywords{Deep Reinforcement Learning, Cooperative Exploration, Continuous
Control Tasks}
         
\newcommand{\BibTeX}{\rm B\kern-.05em{\sc i\kern-.025em b}\kern-.08em\TeX}

\begin{document}

\pagestyle{fancy}
\fancyhead{}

\maketitle 

\section{introduction}
Deep reinforcement learning (DRL)~\cite{sutton_reinforcement_2018}, which utilizes deep neural networks to learn an optimal policy, works brilliantly and has demonstrated beyond human-level performance in solving various challenging sequential decision-making tasks e.g., video games~\cite{mnih2015human,mnih2016asynchronous,chen2021wide,xu2022side,ZHANG2022341}, autonomous driving~\cite{Ye2019automated,chen2022mind}, robotic control tasks~\cite{andrychowicz2017hindsight, lillicrap2015continuous}, etc. In DRL settings, an agent needs to sufficiently explore the environment and collect a set of experiences to obtain an optimal policy. The agent aims to learn an optimal policy to maximize its expected cumulative rewards through trial and error. Therefore, DRL can be regarded as learning from reward feedback from environments. It is essential that during the training phase the agent should be encouraged to explore the environments and gather sufficient reward signals for well-training.

In DRL, exploration has obsessed with a critical problem: submitting solutions too quickly without sufficient exploration, leading to getting stuck at local minima or even complete failure. DRL researchers adopt the neural network to yield the policy with significant feature extraction and expression capabilities in a range of continuous control tasks. Whereas this phenomenal practice has achieved great performance, it is still obsessed with the notorious insufficient exploration problem in continuous control tasks.
Good exploration becomes extremely difficult when the environment is distracting or provides little feedback.
Whereas existing exploration methods remain a problematic drawback -- lacking diversity to explore. The classic exploration methods such as $\epsilon$-Greedy strategies~\cite{mnih2015human} or Gaussian noise~\cite{lillicrap2015continuous} indirectly and implicitly change the style of the exploration. However, in massive situations, diverse styles of exploration are necessary. For instance, in chess games, players should perform different styles of policies (e.g., radical, conservative, etc.) to keep competitive when facing 
various situations; humanoid robots attempt diverse control styles and eventually learn to walk efficiently.

Recent studies enrich diverse styles of policies by constructing relationships between the distribution of policy and trajectories~\cite{achiam2018variational,florensa2017stochastic,sharma2019dynamics,popo,poer}. In Diversity is All You Need (DIAYN)~\cite{eysenbach2018diversity}, authors highlight the diversity of policies plays a significant role in well-training agents. It trains the policy that maximizes the mutual information between the latent variable and the states, then altering the latent variable of the policy network creates multiple policies performing disparately. Although this interesting viewpoint has attracted a spectrum of following works~\cite{achiam2018variational,eysenbach2018diversity,florensa2017stochastic,gregor2016variational,sharma2019dynamics}, the aforementioned methods achieve diverse policies depending on the task in an unsupervised way, resulting in the algorithm performing insufficient generality in different tasks. In fact, the ideal algorithm to implement the diverse policies should be general across a range of tasks, which motivates us to design a \textit{task-oriented} algorithm working towards developing various policies. Eysenbach et al. proposed~\citep{eysenbach2021information} that the learned skills could not construct all the state marginal distributions in the downstream tasks. For task relevance, the mutual information is added as an intrinsic reward for empowerment~\cite{mohamed2015variational,choi2021variational}, but these methods change the original reward function resulting in the performance being extremely sensitive to the trade-off between original and intrinsic rewards.

Our method insights from an interesting phenomenon during the exploration. The critic aims to approximate the accumulated reward by bootstrapping in the actor-critic framework. However, the different critic functions may have great differences even if they approximate the same target due to the function approximation error. For instance, Twin Delayed Deep Deterministic policy gradient (TD3)~\cite{fujimoto2018addressing} presents that two value functions with different initial parameters perform quite differently with  identical targets. It is knotty to measure whether a value function is exact or not, and the gap between these two value functions is termed as \textit{controversy}, which sees a decreasing trend along with the value function updating process. Our intuition can be ascribed that controversy in the value estimation will lead to sub-optimal policies, and these policies have a bias toward message acquisition known as the \texttt{style}.

This paper highlights that controversy can be utilized to encourage policies to yield multiple styles, and encourages exploration for a continuous control task by applying multi-styled policies. Our paper contributes three aspects. (1) We first describe that the estimation bias in double value functions can lead to various exploration \texttt{styles}. (2) This paper proposes the CCEP algorithm, encouraging diverse exploration for environments by cooperation from multi-styled policies. (3) Finally, in CCEP, we design a novel framework, termed as the centralized value function framework, which is updated by experience collected from all the policies and accomplishes the message delivery mechanism between different policies.  
Extensive experiments are conducted on the MuJoCo platform to evaluate the effectiveness of our method. The results reveal that the proposed CCEP approach attains substantial improvements in both average return and sample efficiency on the baseline across selected environments, and the average return of agents trained using CCEP is $6.7\%$ higher than that of the baseline. Besides, CCEP also allows agents to explore more states during the same training time steps as the baseline. Additional analysis indicates that message delivery leading to the cooperative multi-styled policies further enhanced the exploration efficiency by $8.6\%$ compared with that of without cooperation.

We organize the rest of this paper as follows. Section~\ref{sec:preliminaries} briefly explains the concepts of RL. We elaborate on how to perform the CCEP algorithm in Section~\ref{sec:CCEP}. We introduce our experimental settings in Section~\ref{sec:setup}. Section~\ref{sec:result} presents and analyzes results to validate the effectiveness of CCEP, after which section~\ref{sec:related} discusses related works. Finally, we conclude our paper in Section~\ref{sec:conclusion}. The code and documentation are released in the link for validating reproducibility.\footnote{\label{link:open}\url{https://github.com/Jincate/CCEP}}

\begin{figure*}[!t]
    \centering
    \includegraphics[width=0.9\textwidth]{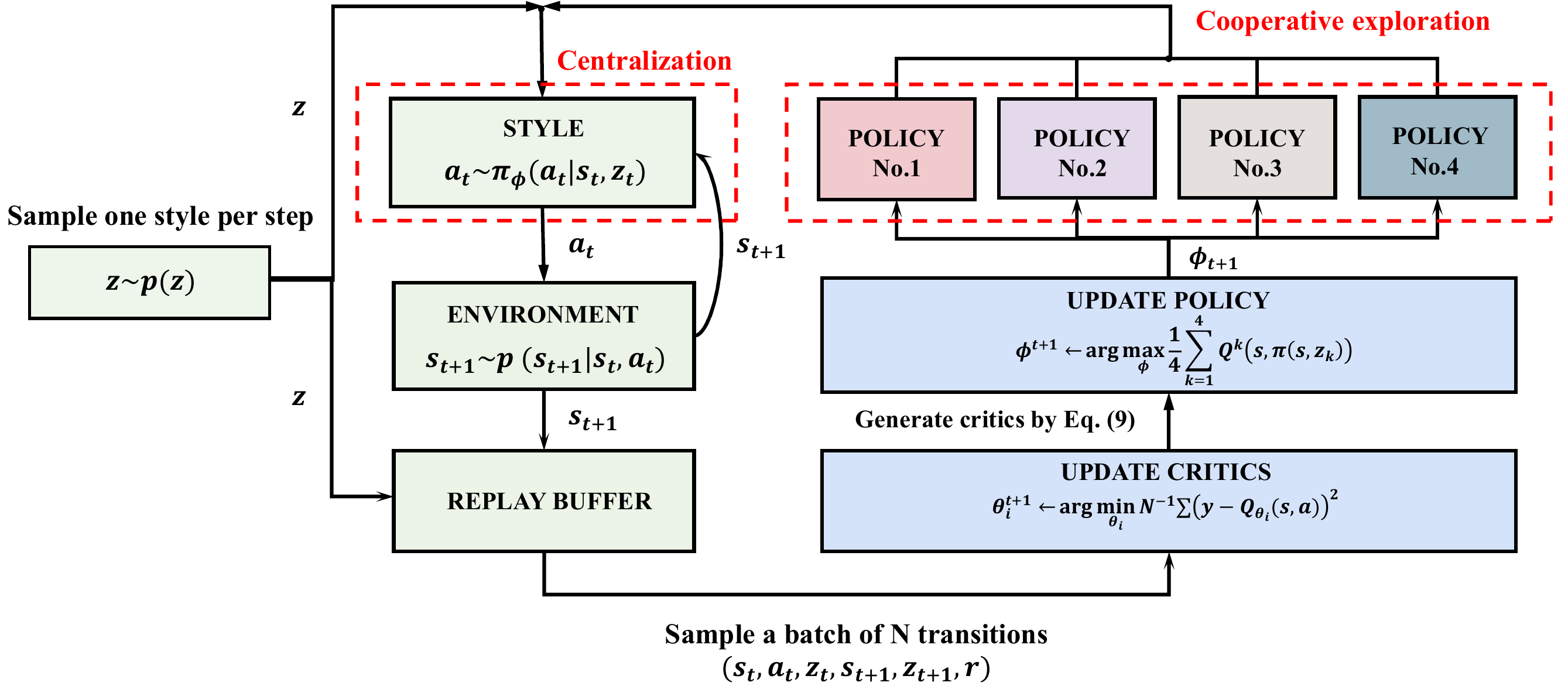}
    \caption{The workflow of CCEP Algorithm. The agent $\pi$ interacts with the environment with diverse style cooperatively and produce the transition $s_t \rightarrow s_{t+1}$. The actor and critic are updated over a mini-batch of the transition samples. A centralized policy with four different styles is learned from the multi-styled critics.}
    \label{fig:workflow}
\end{figure*}
\section{Preliminaries}
\label{sec:preliminaries}
Reinforcement learning (RL) aims at training an agent to tackle the sequential decision problems that can be formalized as a Markov Decision Process (MDP). This process can be defined as a tuple $(\mathcal{S},\mathcal{A},P,r,\gamma)$, where $\mathcal{S}$ is the state space, $\mathcal{A}$ is the action space, $P: \mathcal{S} \times \mathcal{A} \times \mathcal{S} \mapsto[0,1]$ denotes the transition probability, $r(s,a)$ is the reward function $r: \mathcal{S}\times\mathcal{A} \mapsto\mathbb{R}$, 
determining the reward agents will receive in the state $s$ while executing the action $a$. The $\gamma \in (0,1)$ is the discount factor. The return is defined as the discounted accumulated reward.
\begin{equation}
    R=\sum^\infty_{t=0} \gamma^t r(s_t,a_t)
\end{equation}
In the DRL community, developers usually use the neural network parameterized with $\phi$ to indicate the policy $\pi(a|s)$, which inputs an observation and outputs an action. 
The goal of DRL is to solve this MDP process and find the optimal policy $\pi_{\phi^\ast}:\mathcal{S}\mapsto\mathcal{A}$ with parameter $\phi^\ast$ that maximizes the expected accumulated return.

\begin{equation}
    \phi^\ast=\arg\max_{\phi} \mathbb{E}_{a_t\sim \pi_\phi(\cdot|s_t), s_{t+1}\sim P(\cdot|s_t,a_t)}\left[\sum^\infty_{t=0} \gamma^t r(s_t,a_t)\right]
    \label{eq:rl_target}
\end{equation}
David Silver, et al.~\cite{10.5555/3044805.3044850} propose that solving Eq. (\ref{eq:rl_target}) with deterministic policy gradient strategy,
\begin{equation}
\nabla _\phi J(\phi) = \mathbb{E}_{s_{t+1} \sim P(\cdot|s_t,a_t)}\left[\nabla _a Q^\pi(s,a)|_{a=\pi(s)}\nabla _\phi \pi_\phi (s)\right]
\end{equation}
where $Q^\pi(s,a) = \mathbb{E}_{a_t\sim \pi_\phi(\cdot|s_t), s_{t+1}\sim P(\cdot|s_t,a_t)}[R|s,a]$ is known as the value function, indicating how good it is for an agent to pick action $a$ while being in state $s$. To use the gradient-based approach (e.g., Stochastic Gradient Descent~\cite{SGD}) to solve this equation, deep Q-learning uses the neural network to approximate the value function. The value function parameterized with $\theta$ is updated by minimizing the temporal difference (TD) error~\cite{temporal_differences} between the estimated value of the  subsequent state $s'$ and the current state $s$. 
\begin{equation}
\theta^\ast=\arg\min_{\theta} \mathbb{E}\left[r(s,a) + \gamma Q^\pi _\theta(s',a') - Q^\pi _\theta(s,a)\right]^2
\end{equation}
We store the trajectories of the agent exploring the environment in a replay buffer~\cite{replaybuffer} from which sample a random mini-batch of samples, updating the parameters mentioned above.

\section{Centralized Cooperative
Exploration Policy}
\label{sec:CCEP}
This section details technologies of CCEP (Centralized Cooperative
Exploration Policy). We first analyze value estimation bias from function approximation errors and generate multi-styled value functions by encouraging overestimation bias and underestimation bias for the value functions, respectively. To achieve multi-styled exploration, we propose a multi-objective update method for training policy and randomly select one policy to explore at each time step. These historical trajectories during exploration are stored for training a single policy function to achieve cooperative message delivery. 
We denote our policy as $\pi(s,z)$, where $z$ is a one-hot label and represents different policies. In this work, we focus on generating multiple policies with different styles to encourage diverse exploration. We implement our method based on TD3 \cite{fujimoto2018addressing} which maintains double critics and uses the minimum of the critics as the target estimate.

\subsection{Function Approximation Error}
\label{error}
This section shows that there exist approximation errors in the value function optimization and can accumulate to substantial scales. The accumulated approximation error will lead to value estimation bias, which plays a significant role in policy improvement. 

In value-based deep reinforcement, deep neural networks approximate the value functions, and the function approximation error exists correspondingly. One major source of the function approximation error comes from the optimization procedure. In this procedure, stochastic gradient descent, which uses a batch of random samples for gradient update each time, is the mainstream method due to the consideration of computational resources and training efficiency. However, as ~\cite{10.1007/11564096_32} has indicated, a mini-batch gradient update may have unpredictable effects on samples outside the training batch, which leads to the function approximation error. For explanation, we use $e_t$ to represent the approximation error of the value function with the state-action pair$(s_t,a_t)$ as input and approximation error $e_t$ can be modeled as follows:
\begin{equation}
    \label{eq:expand}
    Q_\theta(s_t,a_t) = r(s_t,a_t) + \gamma\mathbb{E}[Q_\theta(s_{t+1},a_{t+1})] - e_t
\end{equation}
Approximation errors influence the value estimation when using the value function as an estimator. The estimation may be skewed towards an overestimation, causing a wrong estimate for a given state. This leads to a problem of an optimal action being chosen but replaced by a sub-optimal action, owing to the overestimation of a sub-optimal action. Thus, the overestimation bias is a common problem in Q-Learning with discrete actions, as we choose the seemly best action $a_{t+1}$ in the target value. Still, there is little chance for the optimal state-action pair to be updated.
\begin{equation}
    \mathbb{E}[\max_{a_{t+1}\in \mathcal{A}}Q(s_{t+1},a_{t+1})]\geq\max_{a_{t+1}\in \mathcal{A}}\mathbb{E}[Q(s_{t+1}, a_{t+1})]
\end{equation}
Mentioned overestimated bias can also occur in continuous control tasks\cite{fujimoto2018addressing}, since the policy approximator always provides the optimal action at the current state based on the value function. 
While this bias can be quite small in an individual update, the bias can be accumulated to a substantial overestimation. Eq.(\ref{eq:expand}) can be expanded as follows:
\begin{equation}
    Q_\theta(s_{t},a_{t}) = \mathbb{E}_{a_t\sim \pi_\phi(\cdot|s_t), s_{t+1}\sim P}\left[\sum^\infty_{t=0} \gamma^t(r(s_t,a_t)-e_t)\right]
\end{equation}
Previous works such as Double Q-learning~\cite{Double_Q_learning} and Double DQN~\cite{Double_DQN} are proposed to alleviate value functions of underestimating. The idea is to maintain two independent estimators in which one is used for estimation while the other is for selecting maximal action. Similarly, as an extension in dealing with continuous control tasks, TD3~\cite{fujimoto2018addressing} reduce the overestimation bias by using double value functions and taking the minimum between the two value functions for an estimation which suffers from underestimation problems as well~\cite{Triplet-Average,Quasi-median}.

\textbf{Does the estimation error influence the performance?} Given a continuous control task, we use $f$ to approximate the true underlying value function $Q^{*}$, which indicates the accumulated reward obtained by acting $a$ before taking optimal policy $\pi^{*}$ at state $s$. $V$ represents the true underlying value function(which is not known during training). $V^*$ and $V^{\pi_f}$ represent the accumulated return obtained by adopting the optimal policy $\pi^{*}$ and $\pi_f$ in state $s$ respectively in which $\pi_f$ is a learned policy by maximizing the value function approximate $f$. 
\begin{lem}
(Performance Gap). The performance gap of the policy between the optimal policy $\pi^{*}$ and the learned policy $\pi_f$ is defined by an infinity norm $\Vert V^*-V^{\pi_f} \Vert_\infty$ and we have
$$\Vert V^*-V^{\pi_f} \Vert_\infty \leq \frac{2\Vert f-Q^{*} \Vert_\infty}{1-\gamma}$$
\label{theorem:1}
\end{lem}
We provide proof detail of Lemma~\ref{theorem:1} in Supplementary~\ref{sup:mis_pro}. This inequality indicates that the performance gap of the policy can be bounded by the estimation error of the value function and accurate value estimate can reduce the upper bound of the performance gap and enhance the performance.\\
\textbf{Do overestimation bias and underestimation bias affect performance in the same way?} 
An empirical study shows that estimation bias may not always be a detrimental problem while both underestimation bias and overestimation bias may improve learning performance which depends on the environment \cite{DBLP:conf/iclr/LanPFW20}. As an example, for an unknown area with high stochasticity, overestimation bias may help to explore the overestimated area but underestimation bias prevents this. However, if these areas of high stochasticity are given low values, the overestimation bias may lead to excess exploration in low-value regions. The fact is that we can not choose the environment and these different circumstances can always occur during exploration. Our method is designed to utilize the difference in exploration behavior brought by estimation bias to encourage multi-styled exploration.

\subsection{Multi-Style Critics: Radical, Conservative}
\label{style-critic}
As mentioned above, function approximation error exists in value functions and can accumulate to substantial scales which have a great influence on the value estimation resulting in overestimation or underestimation bias. Estimation bias has been researched in recent works \cite{fujimoto2018addressing, Triplet-Average,Quasi-median, wd3}. While these works focus on an accurate value estimation and discussed the method to control the estimation bias with the use of multiple value functions for auxiliary, they just choose one of the value functions, which seems to be the most accurate, for policy update neglecting other value functions. However, there is no accurate value function without trial and error. In this section, we show how to utilize the estimation bias and introduce our method for the generalization of multi-styled critics. 

Our intuition is that there are different degrees of estimation bias in double randomly initialized value functions when performing function approximation. However, the estimation bias can be controlled by applying a maximum operator and minimum operator, namely the maximum of the two estimates is relatively overestimated and the minimum of the two estimates is relatively underestimated. Two different estimates raise a controversy about which critic gives the accurate estimate. The best way to resolve the controversy is to follow one of the critics to explore and collect reward messages. While controversy does not always exist because there is only one accurate value, the critics reach an agreement when the state value has been exactly estimated. And the existence of controversy means more exploration is needed.

We start by maintaining double randomly initialized value functions $Q_{\theta_1}$ and $Q_{\theta_2}$ with parameters $\theta_1$ and $\theta_2$ respectively and update the value function with TD3~\cite{fujimoto2018addressing} which takes the minimum between the two value functions as the target value estimate:
\begin{equation}
    y=r+\min_{i=1,2}Q_{\theta_i}(s',a'), a'\sim \pi_\phi
\end{equation}
But the two randomly initialized value functions potentially have different value estimations for a given state-action pair due to the accumulated function approximation error. This difference leads to the result that the two critics may give two different suggestions for the best action choice. While these estimates are relatively overestimated or underestimated, these different criteria for a given state-action pair may lead to a different style of action choice. 
  It is reasonable the estimation is radical if we choose the maximum value of the two to estimate and the estimation is conservative if we choose the minimum value of the two. Thus, we consider four critics:
\begin{equation}
    Q^{j}(s,a)=
    \left\{
    \begin{aligned}
        &Q_{\theta_1}(s,a)               &j=0\\
        &Q_{\theta_2}(s,a)               &j=1\\
        &\max(Q_{\theta_1}(s,a),Q_{\theta_2}(s,a)) &j=2\\
        &\min(Q_{\theta_1}(s,a),Q_{\theta_2}(s,a)) &j=3\\
    \end{aligned}
    \right.
    \label{value_function}
\end{equation}
There exists controversy among these critics, and the controversy can further influence the performance of the policy learned.\\ 

\begin{algorithm}[!t]
\caption{Centralized Cooperative Exploration Policy (CCEP)}
\hspace*{0.05in}\leftline{Initialize critic networks $Q_{\theta_1}$,$Q_{\theta_2}$}
\hspace*{0.05in}\leftline{Initialize actor network $\pi_\phi$ with random parameters
$\theta_1$,$\theta_2$,$\phi$}
\hspace*{0.05in}\leftline{Initialize target networks $\theta'_1\leftarrow \theta_1$,$\theta'_2 \leftarrow \theta_2$,$\phi'\leftarrow \phi$}
\hspace*{0.05in}\leftline{Initialize replay buffer $\mathcal{B}$}
\hspace*{0.05in}\leftline{Initialize number of skills $\mathcal{K}$}
\begin{algorithmic}[1]
\For{$t=1$ to $T$}
\State Sample a skill $z$ from $p(z)$
\State Select action with noise $a\sim \pi_\phi(s,a)+\epsilon, \epsilon \sim \mathcal{N}(0,\sigma)$
\State Observe a reward $r$ and a new state $s'$
\State Store transition tuple $(s,z,a,r,s',z')$ in $\mathcal{B}$
\State Sample mini-batch of $N$ transitions $(s,z,a,r,s',z')$ from $B$
\State $a'\leftarrow \pi_{\phi'}(s',z') + \epsilon, \epsilon \sim
clip(\mathcal{N}(0,\sigma),-c,c)$
\State $y=r+\gamma \min_{i=1,2}Q_{\theta'_i}(s',a')$
\State Update critics: $\theta_i \leftarrow \arg\min_{\theta_i}N^{-1}\sum(y-Q_{\theta_{i}}(s,a))^2$
\If{t mod d}
\State Update policy: 
\State $\nabla_{\phi}J(\phi)=N^{-1}\mathcal{K}^{-1}\sum \nabla_{a} Q^{j}(s,a)|_{a=\pi_{\phi}(s,z)}\nabla_{\phi}\pi_{\phi}(s,z)$
\State Update target networks
\State $\theta'_{i}\leftarrow \tau\theta_i + (1-\tau)\theta'_i$
\State $\phi'_{i}\leftarrow \tau\phi_i + (1-\tau)\phi'_i$
\EndIf
\EndFor
\end{algorithmic}
\label{al:ALG}
\end{algorithm}

\subsection{Opposite Value Functions}
\label{mirror}
\begin{figure}
    \begin{minipage}{0.49\linewidth}
    \centering
    \includegraphics[width=\textwidth]{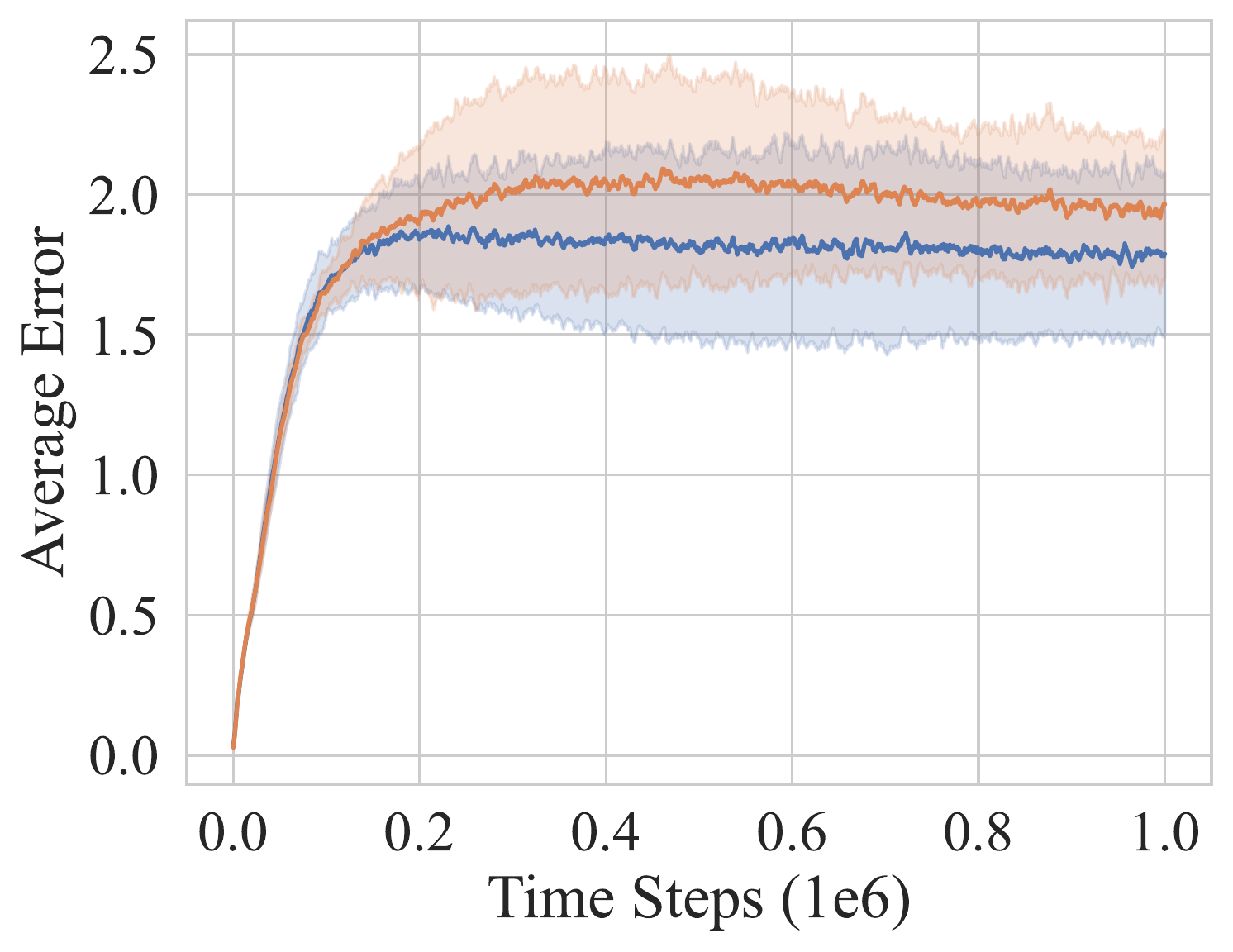}
    \centerline{(a) HalfCheetah-v3}
    \end{minipage}
    \begin{minipage}{0.49\linewidth}
    \centering
    \includegraphics[width=\textwidth]{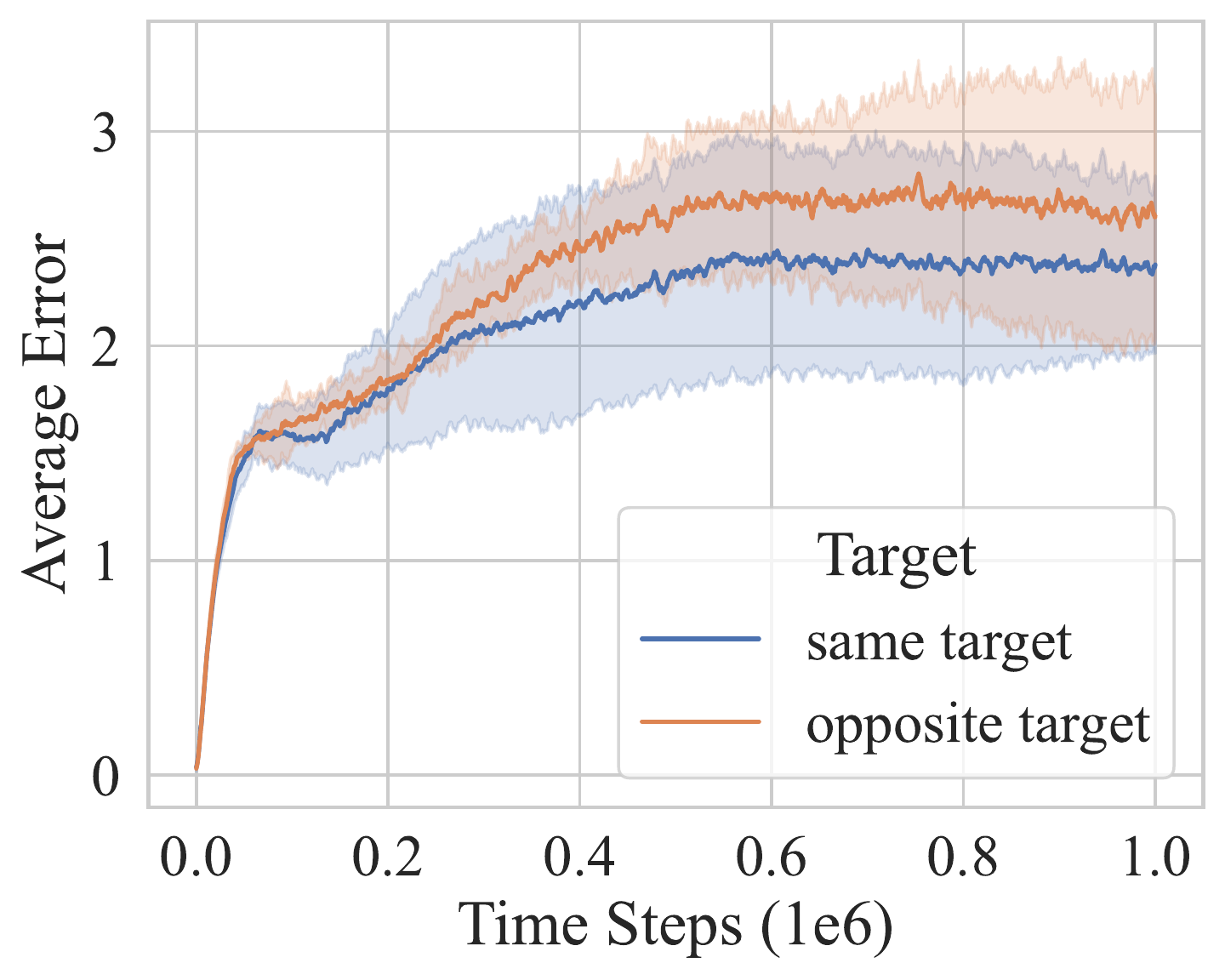}
    \centerline{(b) Walker2d-v3}
    \end{minipage}
    \caption{Measuring the error between double critics given same/opposite targets in TD3 on MuJoCo environments over 1 million time steps}
    \label{fig:error_h}
\end{figure}

This approach for generating diverse styles raises the problem that the value functions may not provide sufficient difference in style when the controversy disappear. This phenomenon is very common when the value function converges. But we don't want this to happen too soon, because we want the value functions to provide more exploration for the policy. 
The controversy exists due to the randomly initialized parameters of the neural networks and the error accumulation. But actually, there is a small probability that the two networks have great similarities, which will lead to double consistent critics. This is not what we want, because consistent critics mean monotonous policy. To avoid this, we try to enlarge the controversy. The solution in this paper is to learn two opposite targets respectively for the two networks, where one of the networks approximates the positive value and another approximates the negative one. This approach is equivalent to adding a factor -1 
to the final layer of either network. We find that the controversy is guaranteed with this simple network structure change. To provide some intuition, we compared the controversy changes after the double value functions learn the opposite target. We express the amount of controversy between the two value functions by the errors of state values in a batch of samples. Figure \ref{fig:error_h} shows the controversy measuring over MuJoCo~\cite{Mujoco} environments in HalfCheetah-v3 and Walker2d-v3. The results show that with the simple network structure change, the controversy is enlarged. But this approach will not influence the value estimation because we just fine-tune the structure.

\subsection{Centralized Cooperation}
\label{policy}
\begin{figure}[!t]
    \centering
    \includegraphics[width=0.48\textwidth]{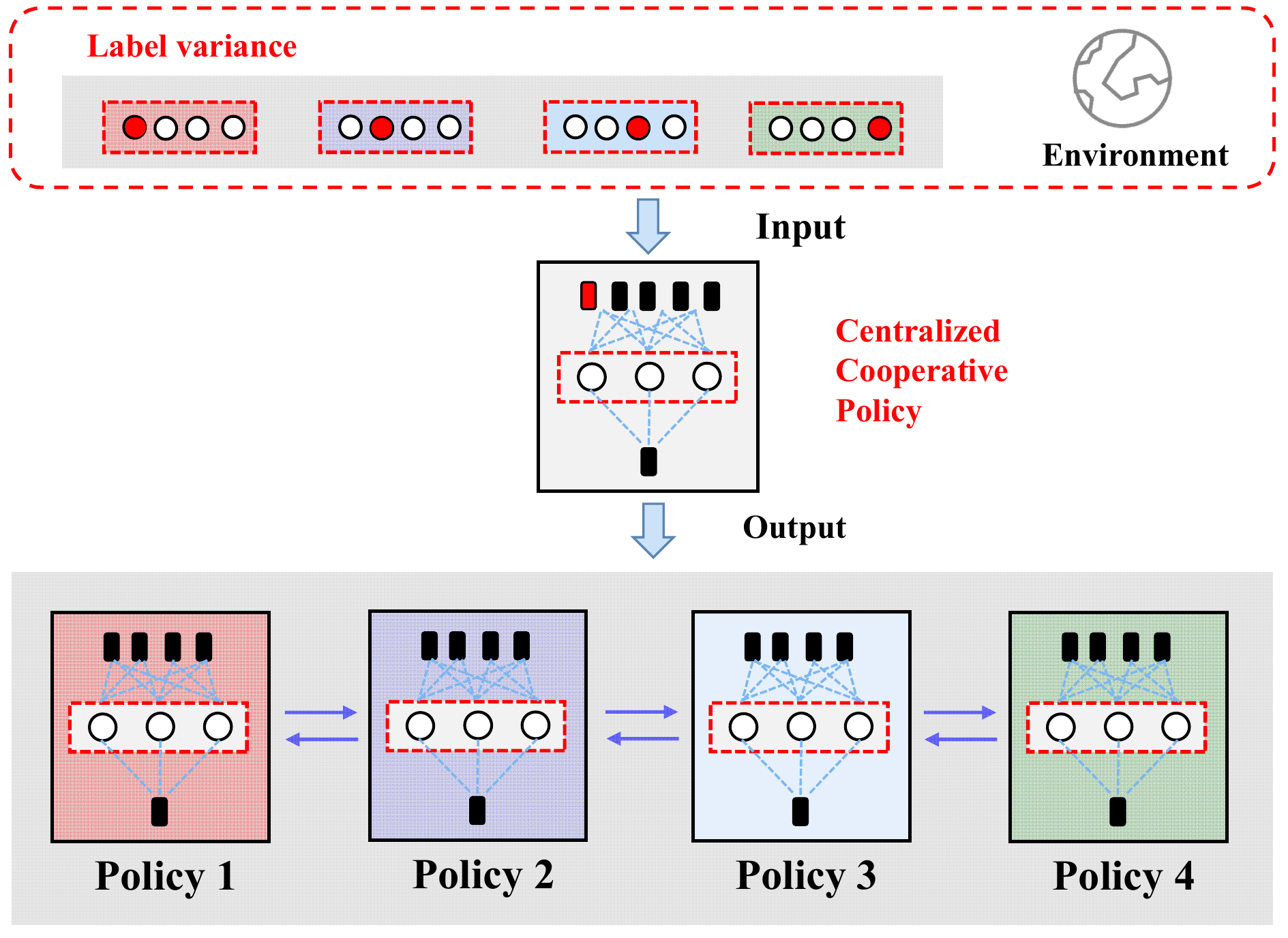}
    \caption{The architecture of our centralized cooperative policy. The agent cooperatively explores the environments by selecting one of the styles at each time step. The style selection process is implemented by sampling latent variable $z$. Policies with diverse styles exchange messages through a centralized network.}
    \label{fig:contralized}
\end{figure}

With four critics, we train a centralized cooperative policy to encourage multi-styled explorations through diverse value estimations. We model this problem as a multi-objective optimization problem. The target is to train multiple policies, with each policy targeting an individual value function. We express the policy function as $\pi(s,z)$, with state $s$ and latent variable $z$ as input. The latent variable $z$, which is a one-hot label in our method, represents different policies. The architecture of our centralized cooperative policy is shown in Figure \ref{fig:contralized}. This idea comes from skill discovery method \cite{eysenbach2018diversity,sharma2019dynamics}, which use the latent variable $z$ to express different skills. And in skill discovery, the target is to maximize the mutual information between latent variable $z$ and some aspects of the trajectories, which is a different target for a different latent variable $z$. Our method encourages diverse styles of policies by different targets as well. Particularly, we sample latent variable $z$ from set $\{0,1,2,3\}$ and encode it in a one-hot label. For a given latent variable $z$, the policy targets $z$-th value functions in Eq.(\ref{value_function}). With different latent variable $z$, the policy shows diverse styles due to the multi-styled targets. We make an experiment showing that there exists different exploration preferences for these policies (Section \ref{sec:style}) In the exploration procedure, we randomly sample latent variable $z$ and make decisions by policy $\pi(s,z)$. This approach enables diverse styles to be applied at each time step. Broadly speaking, our exploration policy has the following characteristics: Multi-styled, Centralized, and Cooperative.

\textbf{Multi-styled.} We train four policies to accomplish the exploration. These policies learn from the corresponding value function $Q^j$:
\begin{equation}
    \pi^{*}_{j} = \arg\max_{\pi}Q^j
\end{equation}
There are four value estimators, in which two of them ($j=0,1$) are normal but different estimators, one ($j=2$) is an overestimated estimator compared to the other ($j=3$) and helps encourage explorations in overestimated actions, the last remaining one ($j=3$) is a conservative estimator and brings more exploitation as illustrated in the previous section. It is appropriate for the policies to perform in a variety of ways given the varied estimators they use (e.g., conservative, radical).

\textbf{Centralized.}  Our policy is a centralized policy because we make use of all the policies learned in each episode. At each time step $t$, we sample one of these policies for exploration. It allows us to generate a variety of trajectories adopting this exploration approach as this centralized policy can generate $4^n$ types of trajectories theoretically for $n$-step exploration, compared to using a single policy that can only generate one. These trajectories are stored as experience and maintain the update for a pair of centralized value functions.

\textbf{Cooperative.} We update the policy cooperatively. With multiple policies learning their respective value functions, ``knowledge'' learned by each policy cannot be shared. Our method is to learn a single network for policies and learn cooperatively~\cite{9891942}. To represent different policies, we feed latent variables $z$ which are one-hot labels as extra input to the network. The policy which inputs latent variable $z$ and state$s$ and outputs action$a$ can be defined as $\pi(s,z)$. We sample $z$ to represent the sampling of different policies. Thus, the policy can be updated by taking deterministic policy gradient.
\begin{equation}
    \nabla_{\phi}J(\phi)=N^{-1}\mathcal{K}^{-1}\sum \nabla_{a} Q^{j}(s,a)|_{a=\pi_{\phi}(s,z)}\nabla_{\phi}\pi_{\phi}(s,z)
\end{equation}
Where $Q^{j}(s,a)$ refer to the multi-styled critics in Eq.(\ref{value_function}), $\mathcal{K}$ is the number of styles which is 4 in this algorithm. The specific algorithm is shown in Algorithm \ref{al:ALG}. 


\section{Experimental Settings}
\label{sec:setup}

\begin{figure}[!t]
    \begin{minipage}{0.49\linewidth}
    \centering
    \includegraphics[width=\textwidth]{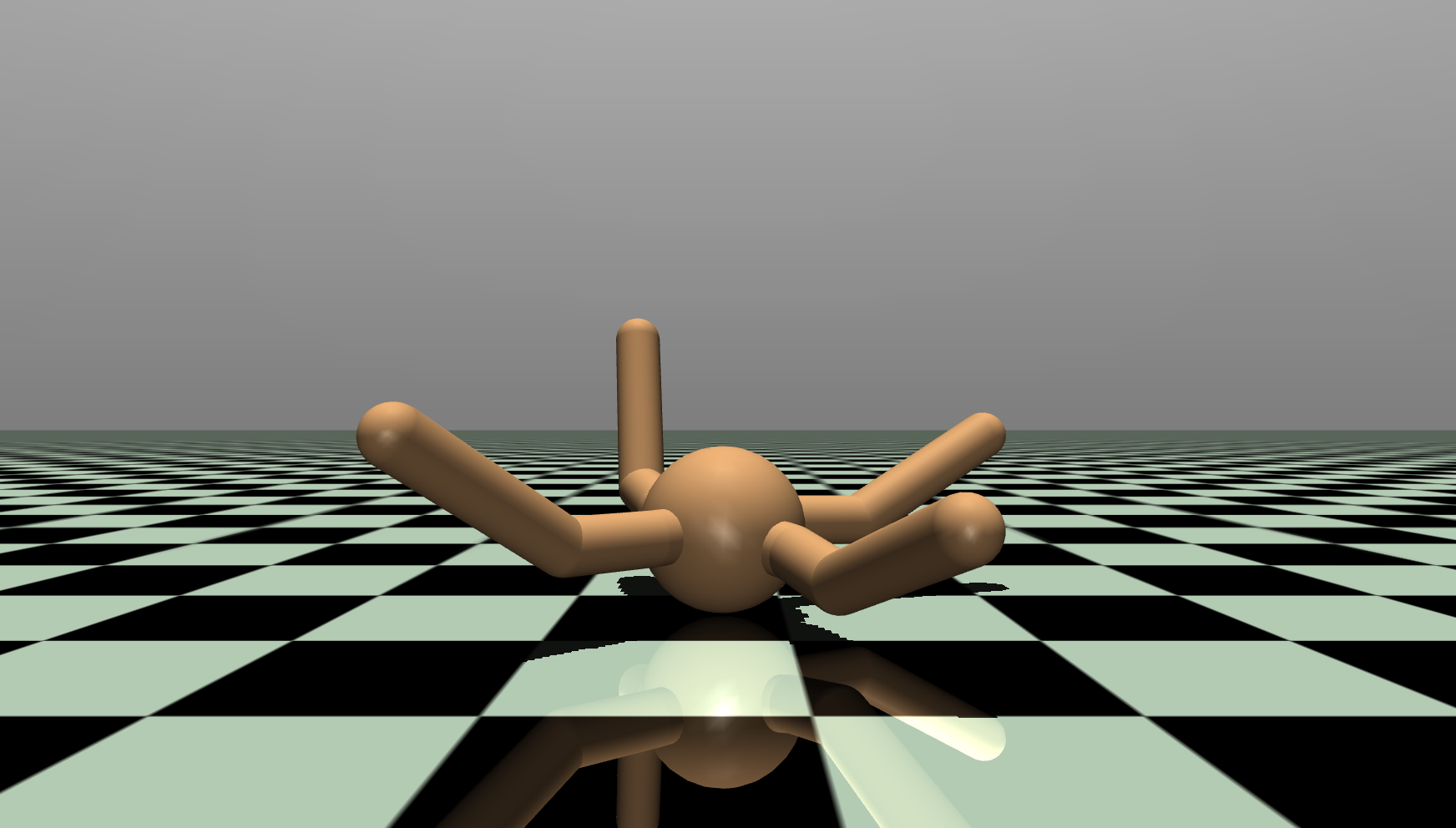}
    \centerline{(a)}
    \end{minipage}
    \begin{minipage}{0.49\linewidth}
    \centering
    \includegraphics[width=\textwidth]{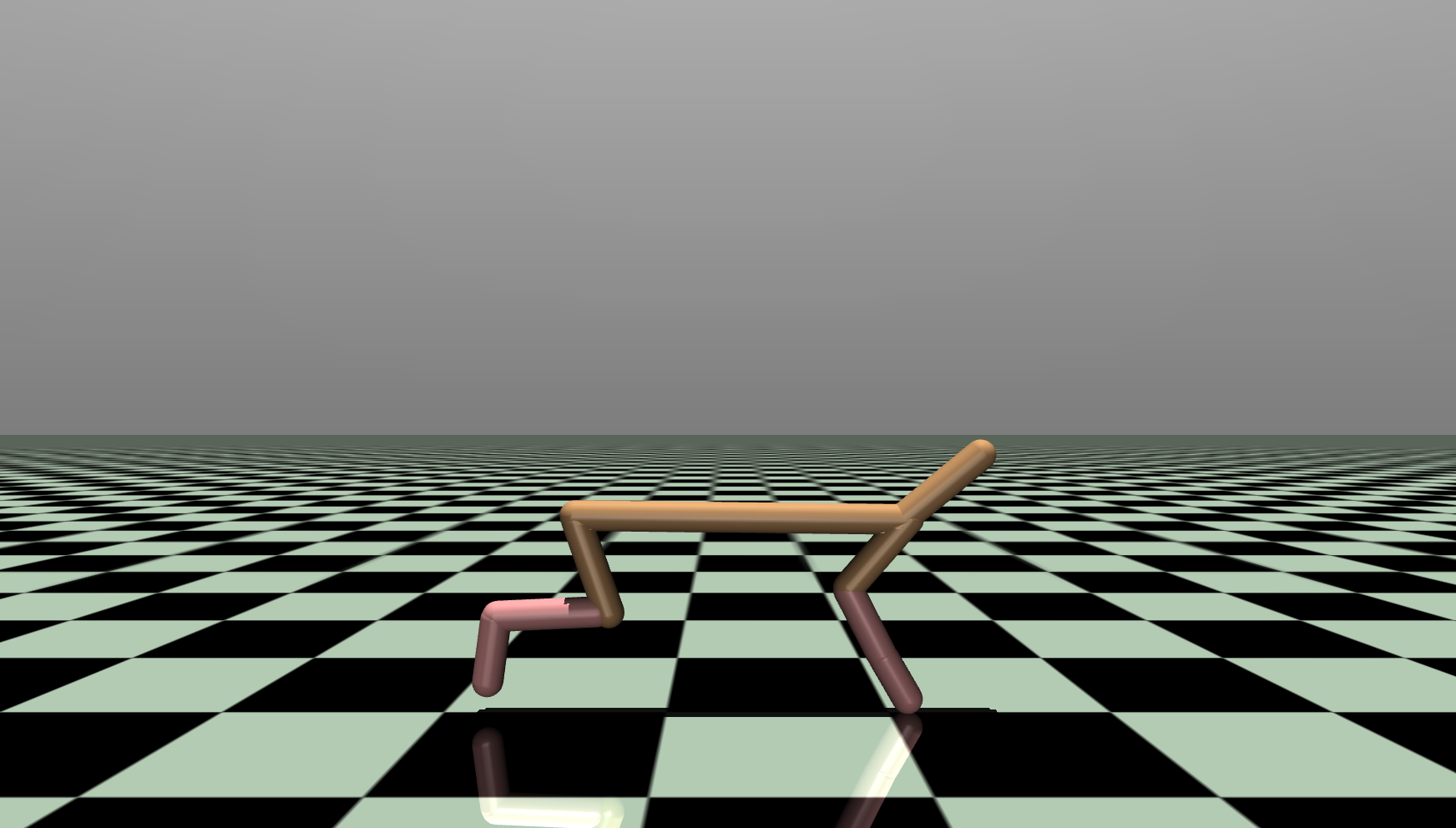}
    \centerline{(b)}
    \end{minipage}
    \begin{minipage}{0.49\linewidth}
    \centering
    \includegraphics[width=\textwidth]{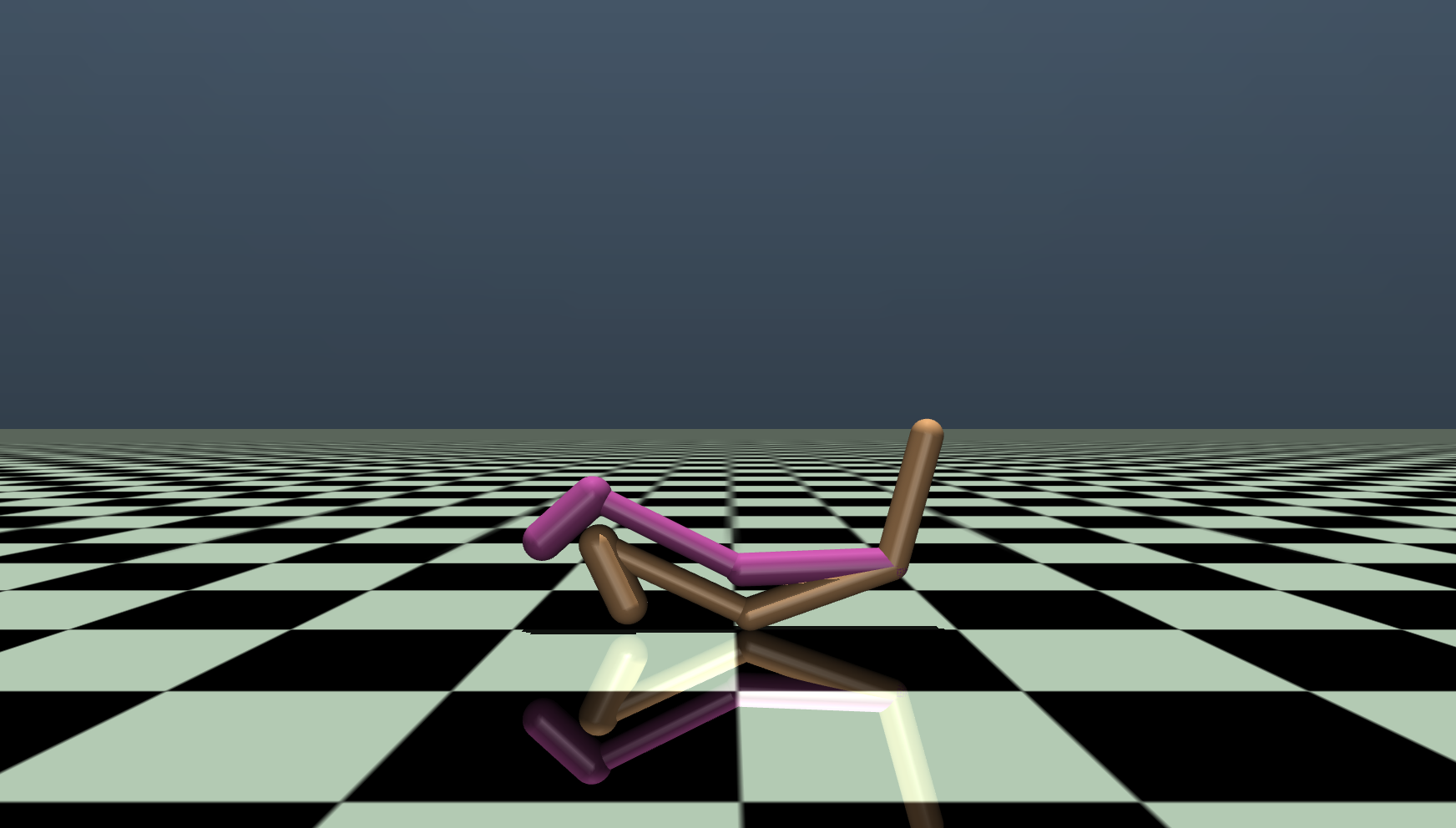}
    \centerline{(c)}
    \end{minipage}
    \begin{minipage}{0.49\linewidth}
    \centering
    \includegraphics[width=\textwidth]{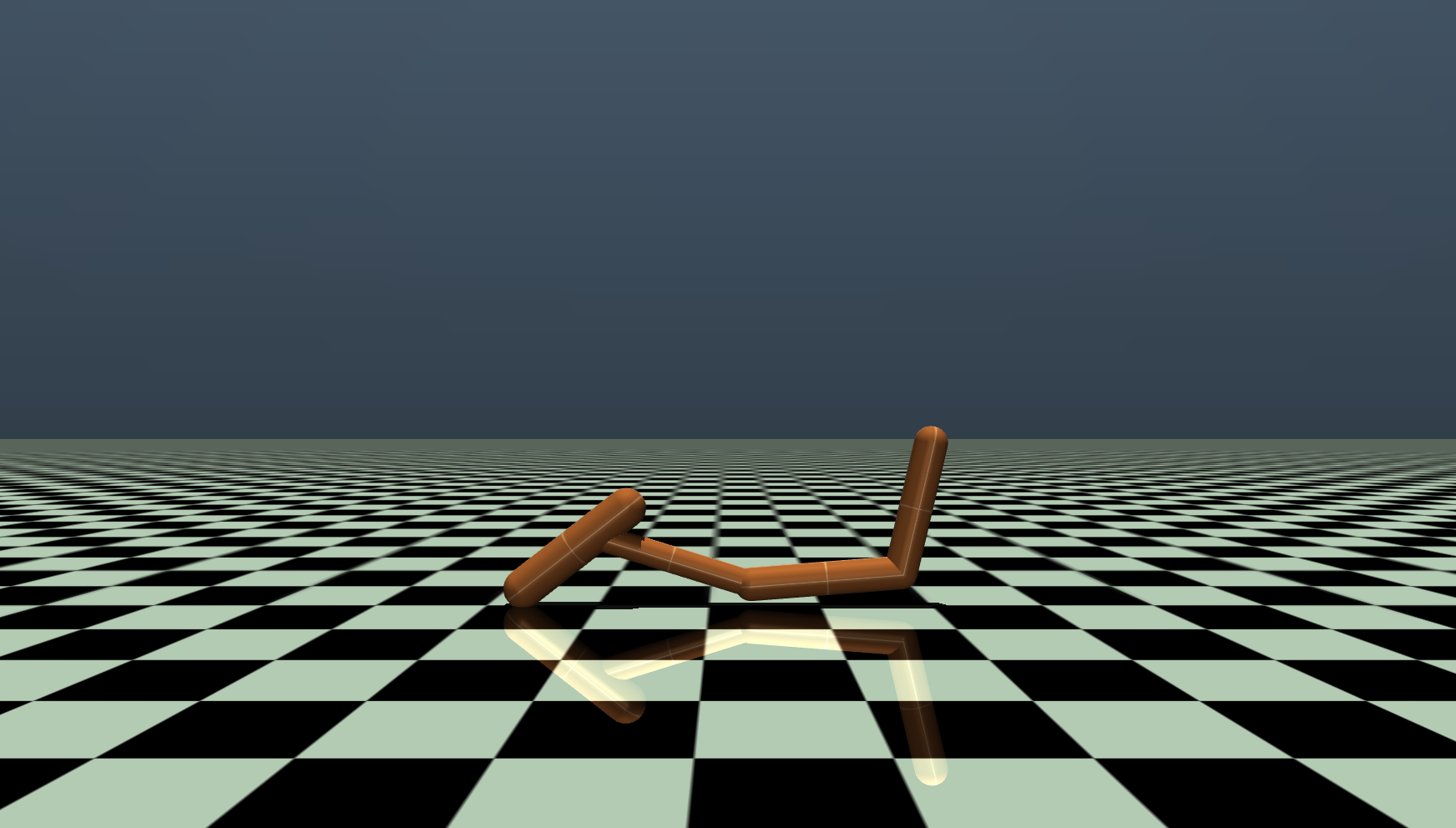}
    \centerline{(d)}
    \end{minipage}
    \centering
    \caption{Screenshots of MuJoCo environments. (a) Ant-v3, (b) HalfCheetah-v3, (c) Walker2d-v3, (d) Hopper-v3}
    \label{fig:Environment}
\end{figure}

To evaluate our method, we test our algorithm on the suit of MujoCo~\cite{Mujoco} continuous control tasks, including HalfCheetah-v3, Hopper-v3, Walker2d-v3, Ant-v3, Pusher-v2 and Humanoid-v3 (the screenshots are presented in Figure~\ref{fig:Environment}). \\
For implementation, our method builds on TD3~\cite{fujimoto2018addressing}, and for comparison, we also establish three-layer feedforward neural networks with 256 hidden nodes per hidden layer for both critics and actors. Particularly, the actor takes state $s$ and latent variable $z$ concatenated as input, where the latent variable $z$ is encoded as one-hot label. At each time step, both networks are trained with a mini-batch of 256 samples. We apply soft updates for target networks as well.

We compared our algorithm against some classic algorithms such as DDPG~\cite{DBLP:conf/iclr/LanPFW20}, which is an efficient off-policy reinforcement learning method for continuous tasks;
PPO~\cite{PPO}, the state-of-the-art policy gradient algorithms; TD3~\cite{fujimoto2018addressing}, which is an extension to DDPG; SAC~\cite{haarnoja2018soft}, which is an entropy-based method with high sample efficiency. Further, we compared our algorithm with the latest algorithm in solving the exploration problems in continuous control tasks such as OAC~\cite{OAC}, which makes improvements on SAC for better exploration. We implement DDPG and PPO by OpenAI’s baselines repository and SAC, TD3, and OAC by the github the author provided. And we use the parameter the author recommend for implementation. The details of the implementation are shown in Supplementary \ref{Implementation_details}.

\begin{figure*}[!t]
    \centering
    \includegraphics[width=0.85\textwidth]{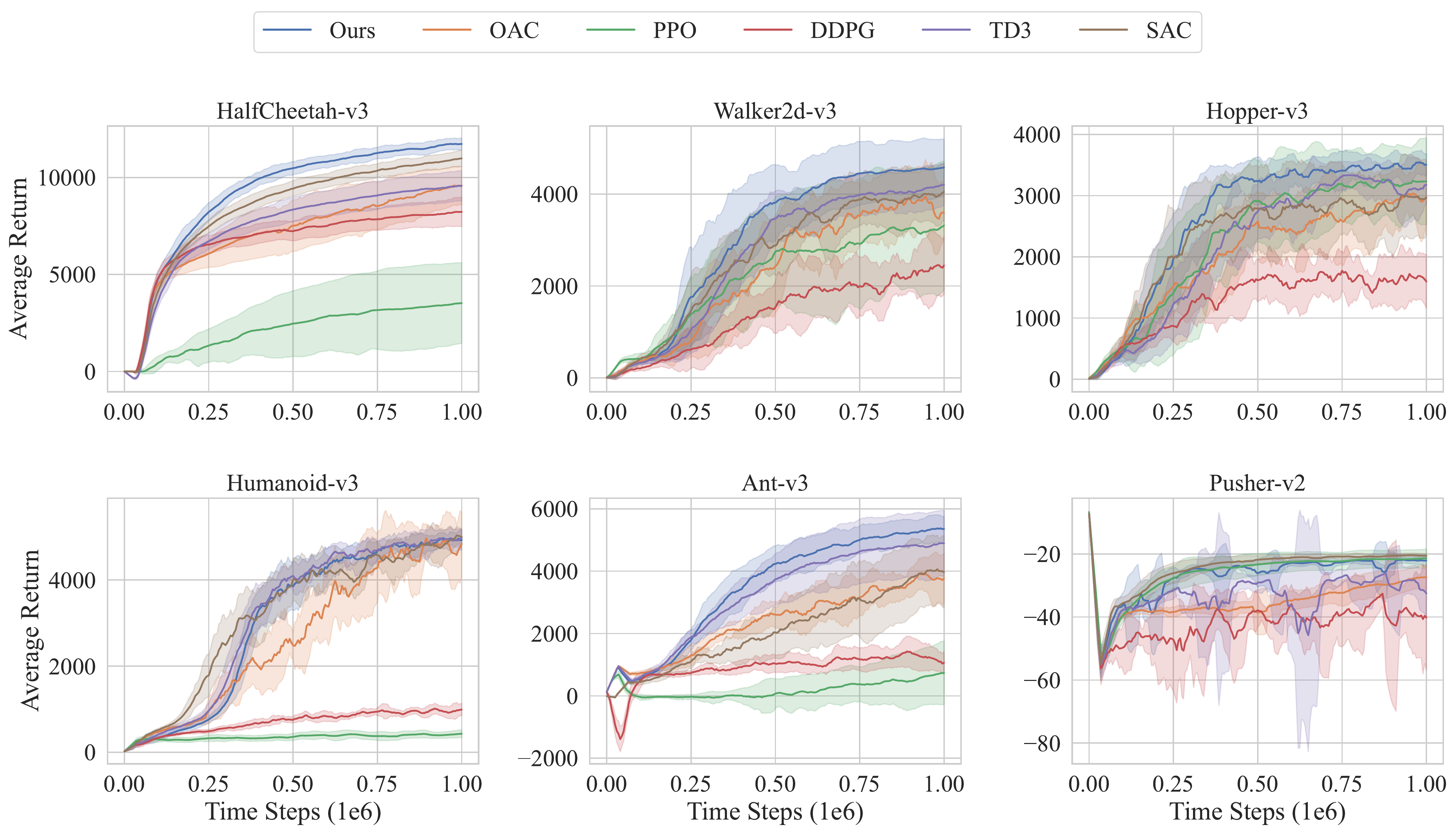}
    \caption{Learning curves for 6 MuJoCo continuous control tasks.For better visualization,the curves are smoothed uniformly. The bolded line represents the average evaluation over 10 seeds. The shaded region represents a standard deviation of the average evaluation over 10 seeds.
}
    \label{fig:results}
\end{figure*}

\section{Experimental Results and Analysis}
\label{sec:result}

\subsection{Evaluation}
To validate the performance of the CCEP algorithm, we evaluate our algorithm in MuJoCo continuous control suites. We perform interactions for 1 million steps in 10 different seeds and evaluate the algorithm over 10 episodes every 5k steps. Our results report the mean scores and standard deviations in the 10 seeds. We show learning curves in Figure \ref{fig:results} and the max average return over 10 trials of 1 million time steps in Table \ref{tab:results}. The learning curves in 1 million time steps show that our algorithm achieves a higher sample efficiency compared with the latest algorithm. Furthermore, the results in the Table \ref{tab:results} indicates that our algorithm shows superior performance. And in HalfCheetah-v3, Walker2d-v3, Hopper-v3, Ant-v3, our algorithm outperforms all the other baselines and achieve significant improvements. While in the Pusher-v2 task, our algorithm show higher stability than that of TD3. For further evaluation, we evaluate our algorithm in the state-based suite PyBullet \cite{coumans2016pybullet} which is considered to be harder than the suite MuJoCo. Our algorithm still shows better performance compared to the baseline algorithms. The corresponding results are shown in Supplementary \ref{sec:add_eval}.

\subsection{Policy Style}
\label{sec:style}
\begin{figure*}
    \centering
    \includegraphics[width=0.80\textwidth]{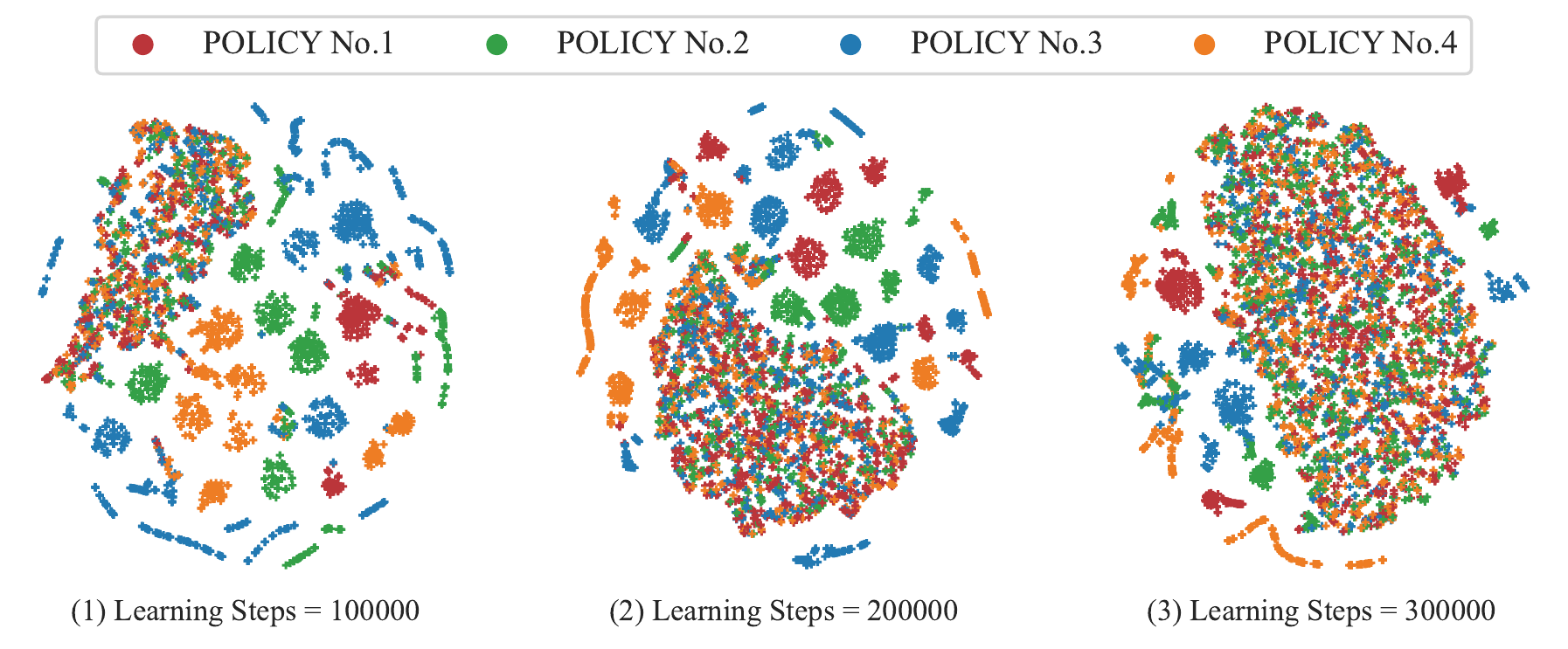}
    \caption{The states visited by each style. For better visualization, the states get dimension reduction by t-SNE. The points with different color represents the states visited by the policy with the style. The distance between points represents the difference between states.}
    \label{fig:StylePartial}
\end{figure*}

To ensure that our proposed CCEP algorithm learns diverse styles, we compared the distribution of explored trajectories when exploring with a single style only. We test the algorithm in Ant-v3 environment over $1e6$ time steps and use the states sampled to represent the trajectories. Figure \ref{fig:StylePartial} shows the states explored by each style at $1e5$, $2e5$ and $3e5$ learning steps, and a more detailed results are shown in Supplementary \ref{sec:sup_style}. We collect the states sampled over 10 episodes with different seeds and apply t-SNE~\cite{TSNE} for better visualization. The results show that while
part of the states can be gathered by all styles which implies a compromise in controversy, there is a considerably large region of states that can only be explored by a unique style of policy. Though different styles, diverse styles come to be in compromise as training process goes on. This phenomenon suggests that CCEP behaves in multi-styled exploration which leads to an exploration preference, and styles come to an agreement with sufficient exploration. Another phenomenal conclusion is that although the style tends to be consistent, new styles are emerging which brings enduring exploration capabilities.

\begin{table}[!t]
    \centering
    \caption{The highest average return over 10 trials of 1 million time steps. The maximum value for each task is bolded.}
    \begin{tabular}{lcccccc}
    \toprule
    Environment & Ours & OAC & SAC & TD3 & DDPG & PPO \\
    \midrule
    HalfCheetah & \textbf{11945} & 9921 & 11129 & 9758 & 8469 & 3681 \\
    Hopper      & \textbf{3636}  & 3364 & 3357 & 3479 & 2709 & 3365 \\
    Walker2d    & \textbf{4706}  & 4458 & 4349 & 4229 & 3669 & 3668 \\
    Ant         & \textbf{5630}  & 4519 & 5084 & 5142 & 1808 & 909 \\
    Pusher      & -21     & -25 & \textbf{-20}  & -25  & -29  & -21  \\
    Humanoid    & 5325 & \textbf{5747} & 5523 & 5356 & 1728 & 586  \\
    \bottomrule
    \end{tabular}
    \label{tab:results}
\end{table}

\subsection{Measuring Exploration Ability}

\begin{figure*}
    \centering
    \includegraphics[width=0.80\textwidth]{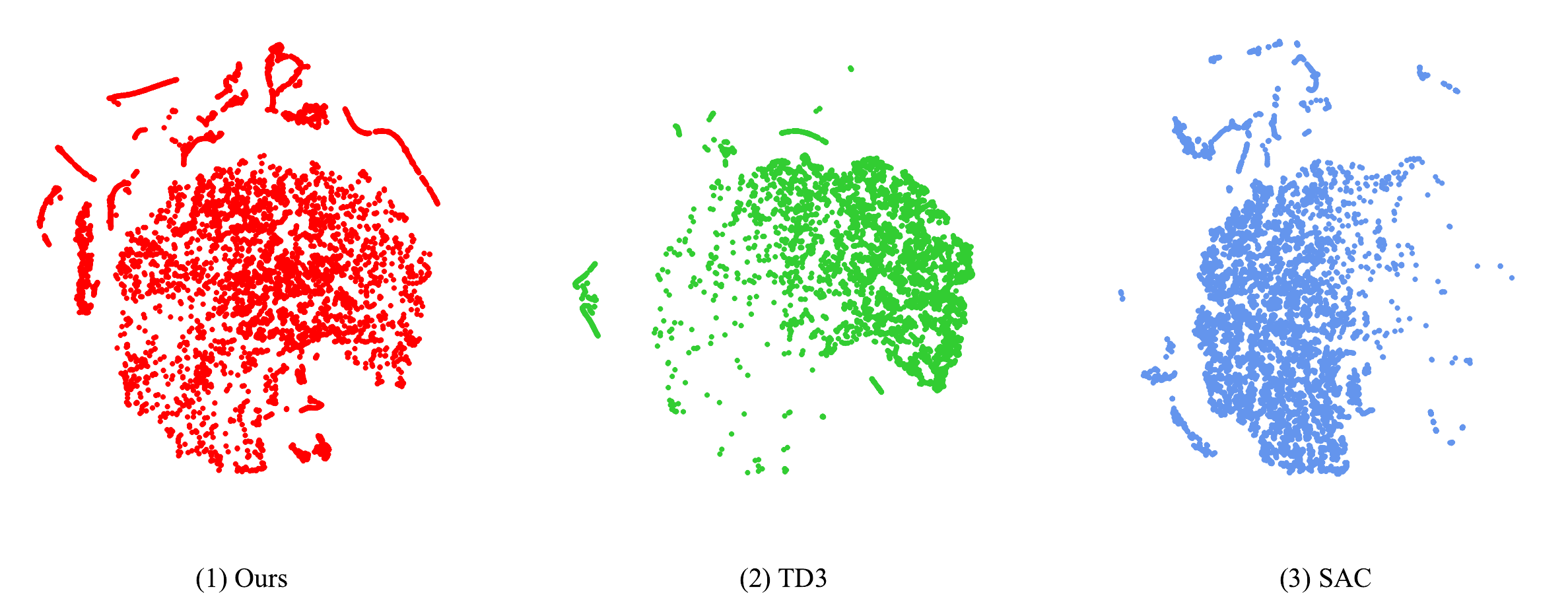}
    \caption{Measuring the exploration region. Comparison of exploration capabilities of TD3 (green), SAC (blue) and Ours (red). The points represent region explored by each method in 10 episodes. All the states get dimension reduction by the same t-SNE transformation for better visualization.}
    \label{fig:region}
\end{figure*}

The critical problem of our proposed method is whether we achieve higher sample efficiency. Although the learning curves (Figure \ref{fig:results}) gives considerably convincing results, a more intuitive result has been given in Figure \ref{fig:region}. We compared the exploration of CCEP with that of TD3 and SAC (which achieve the trade-off between exploration and exploitation by entropy regularization.) over 10 episodes with different seeds (Figure \ref{fig:region}). For a fair comparison, these methods are trained in Ant-v3 with the same seed at half of the training process. In order to get reliable results, the states explored are gathered in 10 episodes with different seeds. We still apply the same t-SNE~\cite{TSNE} transformation to the states explored by all of the algorithms for better visualization. While there are great differences between the states explored by TD3 (green) and SAC (blue), the result shows that our algorithm (red) explores a wider range of states which even covers that TD3 and SAC explored.

\begin{table}[!t]
    \centering
    \caption{Max Average Return over 5 trials of 1 million time steps, comparing ablation over cooperation for message delivery. The maximum value for each task is bolded. }
    \begin{tabular}{lcccc}
    \toprule
    Method & HCheetah & Hopper & Walker2d & Ant\\
    \midrule
    CCEP & \textbf{11969} & \textbf{3672} & \textbf{4789} & \textbf{5488}\\
    CCEP-Cooperation & 11384  & 3583  &  4087 & 4907 \\ 
    TD3 & 9792  & 3531  & 4190 & 4810 \\
    \bottomrule
    \end{tabular}
    \label{tab:ablation}
\end{table}

\subsection{Ablation Study}
\label{sec:ablation}
We perform an ablation study to understand the contribution of the cooperation between policies for message delivery. The results are shown in Table \ref{tab:ablation} where we compare the performance of training policies by removing policy cooperation and training them separately. We perform interactions for 1 million time steps for each method. The results show that without cooperation, the policy network not only trains 4 times more network parameters but also fails to reduce performance. And this performance degradation is even more pronounced on Walker2d. Additional learning curves can be found in Supplementary \ref{sec:add_abla}.
 
\section{Related Work}
\label{sec:related}
This section discusses several methods proposed recently for improving the exploration of deep reinforcement learning. \\
A range of works take an effort in encouraging explorations with the use of randomness over model parameters~\cite{NIPS2011_e53a0a29}. 
Another prevalent series of works propose to enhance exploration by simultaneously maximizing the expected return and entropy of the policy~\cite{haarnoja2018soft, ziebart2008maximum,Rawlik-RSS-12, 10.5555/3294996.3295018,soft_q_learn}. Whereas, these methods do not provide heuristic knowledge to guide the exploration, which can be considered to be insufficient and time-consuming.

To achieve effective exploration, the {\em curiosity mechanism}~\cite{Raileanu2020RIDE,gongchen2022curiosity} has been proposed in recent works, e.g., the counted-based approaches~\cite{machado2020count} which quantify the ``novelty'' of a state by the times visited. However, these methods maintain the state-action visitation counts which make it challenging in solving high-dimensional or continuous tasks. Other works rely on errors in predicting dynamics, which have been used to address the difficulties in complex environments~\cite{pathak2017curiosity,burda2018exploration,Raileanu2020RIDE}. Though the Intrinsic Curiosity Module (ICM)~\cite{pathak2017curiosity} maintains a predictor on state transitions and considers the prediction error as an {\em intrinsic reward}, 
Random Network Distillation (RND)~\cite{burda2018exploration} utilizes the prediction errors of networks trained on historical trajectories to quantify the novelty of states, which is effective and easy to implement in real applications.

Another direction in previous work is to study exploration in hierarchical reinforcement learning (HRL)~\cite{sutton1999between, bacon2017option}. These methods are insight from the fact that developers prefer to divide the comprehensive and knotty problems into several solvable sub-problems. There are some further studies on hierarchy in terms of tasks, representative of which are goal-based reinforcement learning and skill discovery. The similarity of these approaches is that they both identify different policies by utilizing latent variables. 
In goal-based RL, the latent variables are defined by the policy's goal, which aims to complete several sub-goals and accomplish the whole task. These methods introduce prior human knowledge, causing them to work brilliantly on some tasks but fail when unaware of human knowledge. Despite our method also introducing latent variables to represent different styles of policies, all the policies share the same objective, nevertheless differing in the 
road to reach the destination.

Skill discovery methods, which adopt the latent variable to represent the skill learned from the policy, introduce mutual information to organize relationships between the latent variable $z$ and some aspects of the trajectories to acquire diverse skills (also known as \textit{style})~\cite{achiam2018variational,eysenbach2018diversity,florensa2017stochastic,gregor2016variational,sharma2019dynamics}. Nevertheless, these methods train the policy in an unsupervised way~\cite{eysenbach2018diversity,sharma2019dynamics,florensa2017stochastic}, suggesting that the skills trained are unaware of task-driven, and they cannot represent the optimal policies when adapted to downstream tasks illustrated in ~\cite{eysenbach2021information}. Our method avoids this issue because we train the policy task-oriented and demonstrate the benefit brought by the attention of these policies to the state value making them differ considerably in exploration style. For task relevance, some related works that learn skills by jointly learning a set of skills and a meta-controller~\cite{dayan1992feudal,florensa2017stochastic,krishnan2017ddco,bacon2017option,heess2016learning,frans2017meta}. The options of the meta-controller control different attentions of each policy. However, these methods usually choose the best option to explore and rarely execute sub-optimal options, 
leading to the drawback -- the algorithm tends to ignore sub-optimal actions that maybe fail in most states but are effective in a few critical scenarios. Our proposed approach randomly selects different styles of policies for directed cooperative exploration, which are improved accordingly with the value function and produce different styles due to differences in attention.

\section{Conclusion}
\label{sec:conclusion}

In the value-based method, value estimation bias has been a common problem. While different estimation bias in double value functions lead to value function controversy, the controversy can be utilized to encourage policies to yield multiple styles. In this paper, we aim at encouraging explorations by multi-styled policies. We start by analysis on estimation bias during the value function training process and its effect on the exploration. We then encourage this controversy between the value functions and generate four critics for producing multi-styled policies. Finally, we apply these policies with diverse styles for centralized cooperative exploration which perform superior sample efficiency in the test environment. 
Though there are a lot of works focusing on reducing the estimation bias for an accurate value estimation, few works try to utilize these inevitable errors to make improvements. Our results show that it is also an option to use the errors to encourage explorations. For future work, it is an exciting avenue for focusing on more expressive policy styles. A style that can be represented as a continuous distribution may be more efficient and more expressive.

\balance
\bibliographystyle{ACM-Reference-Format} 
\bibliography{AAMAS_2023}

\clearpage

\begin{center}
    \huge{Supplementary Materials}
\end{center}

\setcounter{section}{0}
\setcounter{equation}{0}
\renewcommand\thesection{\Alph{section}}
\section{Proof of Lemma~\ref{theorem:1}}
\label{sup:mis_pro}

(\textbf{Lemma 1})(Performance Gap). Let $V^*$ be the ground truth state value in Bellman value iterations, $Q^*$ be the ground truth state action value, $V^{\pi_{f}}$ be the state value when applying learned policy $\pi_f$, $f$ be the value function approximator. The performance gap of the policy between the optimal policy $\pi^{*}$ and the learned policy $\pi_f$ is defined by an infinity norm $\Vert V^*-V^{\pi_f} \Vert_\infty$ and we have\\
\centerline{$\Vert V^*-V^{\pi_f} \Vert_\infty \leq \frac{2\Vert f-Q^{*} \Vert_\infty}{1-\gamma}$}
\leftline{\textit{Proof.} For any $s\in\mathcal{S}$}
\begin{equation}
    \begin{aligned}
        V^{*}(s)-V^{\pi_{f}}(s)=&Q^*(s,\pi^{*}(s))-Q^{*}(s,\pi_{f}(s))\\
        &+Q^{*}(s,\pi_{f}(s))-Q^{*}(s,\pi_{f}(s))\\
        \leq& Q^{*}(s,\pi^{*}(s))-f(s,\pi^{*}(s))\\
        &+f(s,\pi_{f}(s))-Q^{*}(s,\pi_{f}(s))\\
        &+\gamma \mathbb{E}_{s'\sim P(s,\pi_{f}(s))}[V^{*}(s')-V^{\pi_{f}}(s')]\\
        \leq& 2\Vert f-Q^{*} \Vert_\infty + \gamma \Vert V^{*}-V^{\pi_f} \Vert_\infty
    \end{aligned}\nonumber
\end{equation}
\section{Experimental Details} 
\label{Implementation_details}
\subsection{Environments}
We evaluate the performance of CCEP on environments from MujoCo Control Suite~\cite{Mujoco}which can be listed as HalfCheetah-v3, Ant-v3, Walker2d-v3, Humanoid-v3, Hopper-v3, and Pusher-v2, and the specific parameters of these environments are listed in Table \ref{tab:environment}. We use the publicly available environments without any modification.

\subsection{Implementation and Hyper-parameters}

\label{supp:imple_hyper}
Here, we describe certain implementation details of CCEP. For our implementation of CCEP, we follows a standard actor-critic framework. 

\subsection{Soft Actor-Critic Implementation Details}
For implementation of SAC, we use the code the author provided and use the parameters the author recommended. We use a single Gaussian distribution and use the environment-dependent reward scaling as described by the authors. For a fair comparison, we apply the version of soft target update and train one iteration per time step. We use the reward scales as the author recommended (except for Pusher-v2 which is not mentioned by the author in the article). Considering that there are similar action dimensions between Pusher-v2 and HalfCheetah-v3, we set the same reward scale for Pusher-v2. The specific reward scales for each environment is shown in Table \ref{tab:reward_scale}.
\begin{table}[H]
    \centering
    \caption{Environment Specific Parameters}
    \begin{tabular}{l|c|c}
    \toprule
        Environment      & State Dimensions & Action Dimensions \\
    \hline
        Ant-v3           & 111              & 8                 \\
        HalfCheetah-v3   & 17               & 6                 \\
        Hopper-v3        & 11               & 3                 \\
        Humanoid-v3      & 376              & 17                \\
        Pusher-v2        & 23               & 7                 \\
        Walker2d-v3      & 17               & 6                 \\
    \bottomrule
    \end{tabular}
    \label{tab:environment}
\end{table}
\begin{table}[H]
    \centering
    \caption{SAC Environment Specific Parameters}
    \begin{tabular}{c|c}
    \toprule
        Environment     & Reward Scale  \\
        \hline
        Ant-v3          &       5       \\
        HalfCheetah-v3  &       5       \\
        Hopper-v3       &       5       \\
        Humanoid-v3     &       20      \\
        Pusher-v2       &       5       \\
        Walker2d        &       5       \\
    \bottomrule
    \end{tabular}
    \label{tab:reward_scale}
\end{table}
\subsection{Optimistic Actor-Critic Implementation Details}
The implementation of OAC is mainly based on the open source code. We set the hyper-parameters the same as OAC used in MuJoCo which is listed in Table \ref{tab:oac_param}. And for fair comparison, we train with 1 training gradient per environment step. We use the same reward scales as SAC, listed in Table \ref{tab:reward_scale}. 
\begin{table}[H]
    \centering
    \caption{SAC Environment Specific Parameters}
    \begin{tabular}{c|c}
    \toprule
        Parameter           &     Value  \\
        \hline
        shift multiplier $\sqrt{2\delta}$   &       6.86       \\
        $\beta_{UB}$        &       4.66       \\
        $\beta_{LB}$        &       -3.65       \\
    
    \bottomrule
    \end{tabular}
    \label{tab:oac_param}
\end{table}
\subsection{Reproducing Other Baselines}
For reproduction of TD3, we use the official implementation ( https://github.com/sfujim/TD3). For reproduction of DDPG and PPO we use OpenAI’s baselines repository and apply default hyper-parameters.
\begin{table}[H]
    \centering
    \caption{CCEP Parameters settings}
    \begin{tabular}{l|c}
    \toprule
        Parameter & Value \\
        \hline
        Exploration policy & $\mathcal{N}(0,0.1), z\sim p(z)$\\
        Number of policy & 4 \\
        Variance of exploration noise & 0.2 \\
        Random starting exploration time steps & $2.5\times10^4$\\
         Optimizer & Adam\cite{DBLP:conf/iclr/LanPFW20}\\
        Learning rate for actor & $3\times10^{-4}$\\ 
        Learning rate for critic & $3\times10^{-4}$\\
        Replay buffer size & $1\times10^6$\\
        Batch size & 256\\
        Discount $(\gamma)$ & 0.99\\
        Number of hidden layers & 2\\
        Number of hidden units per layer & 256\\
        Activation function & ReLU\\
        Iterations per time step & 1\\
        Target smoothing coefficient $(\eta)$ & $5\times10^{-3}$ \\
        Variance of target policy smoothing & 0.2 \\
        Noise clip range & $[-0.5,0.5]$\\
        Target critic update interval & $2$\\
    \bottomrule    
    \end{tabular}
    
    \label{tab:parameter_settings}
\end{table}

\begin{figure*}
    \centering
    \includegraphics[width=0.98\textwidth]{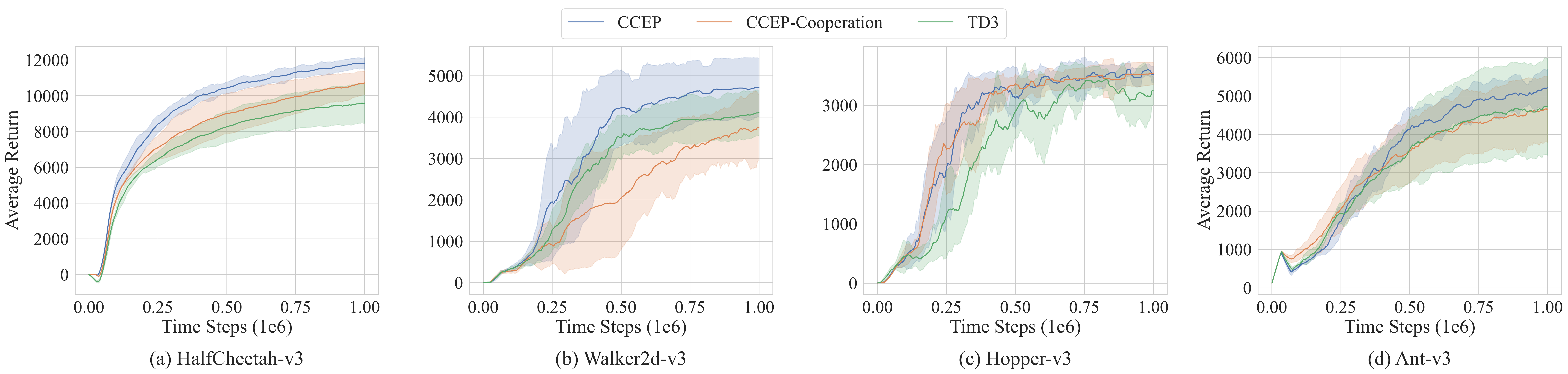}
    \caption{Ablation over the use of cooperation. Comparison of CCEP, TD3 and the subtraction of cooperation (CCEP-cooperation). }
    \label{fig:ablation}
\end{figure*}
\begin{figure*}[hbt]
    \centering
    \includegraphics[width=0.98\textwidth]{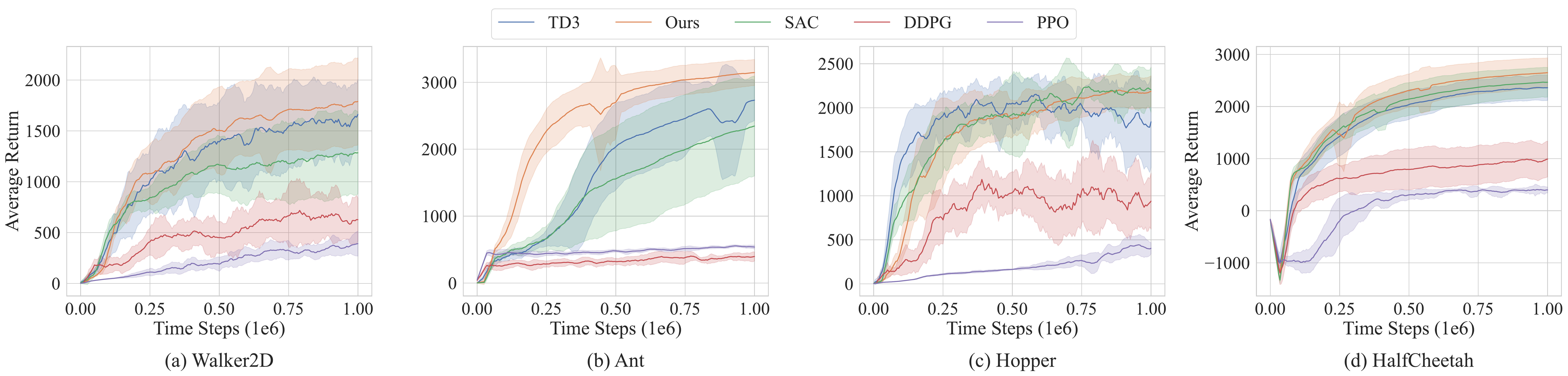}
    \caption{Learning curves for 4 PyBullet continuous control tasks. For better visualization, the curves are smoothed uniformly. The bolded line represents the average evaluation over 10 seeds. The shaded region represents the standard deviation of the average evaluation over 10 seeds.}
    \label{fig:pybullet}
\end{figure*}
\begin{table*}
    \centering
    \caption{Evaluation in PyBullet control suite. The highest average return over 10 trials of 1 million time steps. The maximum value for each task is bolded.}
    \begin{tabular}{lccccc}
    \toprule
    Pybullet Environment & Ours & SAC & TD3 & DDPG & PPO \\
    \midrule
    HalfCheetah & $\boldsymbol{2670\pm275}$ & $2494\pm266$ & $2415\pm236$ & $1120\pm373$ & $465\pm30$ \\
    Hopper      & $\boldsymbol{2254\pm186}$ & $2167\pm323$ & $1860\pm288$ & $1762\pm368$ & $623\pm131$ \\
    Walker2d    & $\boldsymbol{1829\pm418}$ & $1369\pm408$ & $1676\pm342$ &  $929\pm345$ & $509\pm106$ \\
    Ant         & $\boldsymbol{3175\pm184}$ & $2423\pm680$ & $2711\pm253$ &  $483\pm70$ & $578\pm19$ \\
    \bottomrule
    \end{tabular}
    \label{tab:bullet_results}
\end{table*}

\section{Additional Experiments}
\subsection{Additional Evaluation}
\label{sec:add_eval}
For an additional Evaluation, We conduct experiments on the state-based PyBullet \cite{coumans2016pybullet} suite which is based on the well-known open-source physics engine bullet and is packaged as a Python module for robot simulation and learning. The suite of Pybullet is considered to be a harder environment than MuJoCo \cite{Mujoco}.  We choose TD3 \cite{fujimoto2018addressing}, SAC \cite{haarnoja2018soft}, PPO \cite{PPO}, DDPG \cite{lillicrap2015continuous} as our baselines due to their superior performance. We perform interactions for 1 million steps in 10 different seeds and evaluate the
algorithm over 10 episodes every 5k steps. We evaluate our algorithm in HalfCheetah, Hopper, Walker2d and ant in the suite of pybullet. Our results report the
mean scores and standard deviations in the 10 seeds. We show the learning curves in Figure \ref{fig:pybullet} and the max average return over 10 trials in Table \ref{tab:bullet_results}.

\subsection{Additional Ablation Results}
\label{sec:add_abla}
We compare the learning curves of CCEP, TD3 and the subtraction of cooperation (CCEP-cooperation) for better understanding the contribution of policy cooperation (Section \ref{sec:ablation}). We perform interactions for 1 million steps in 10 different seeds and evaluate over 10 episodes every 5k steps. Our results report the
mean scores and standard deviations in the 10 seeds. We show the learning curves in Figure \ref{fig:ablation}

\subsection{Supplementary Results}
\label{sec:sup_style}
\begin{figure*}[htbp]
    \centering
    \includegraphics[width=0.9\textwidth]{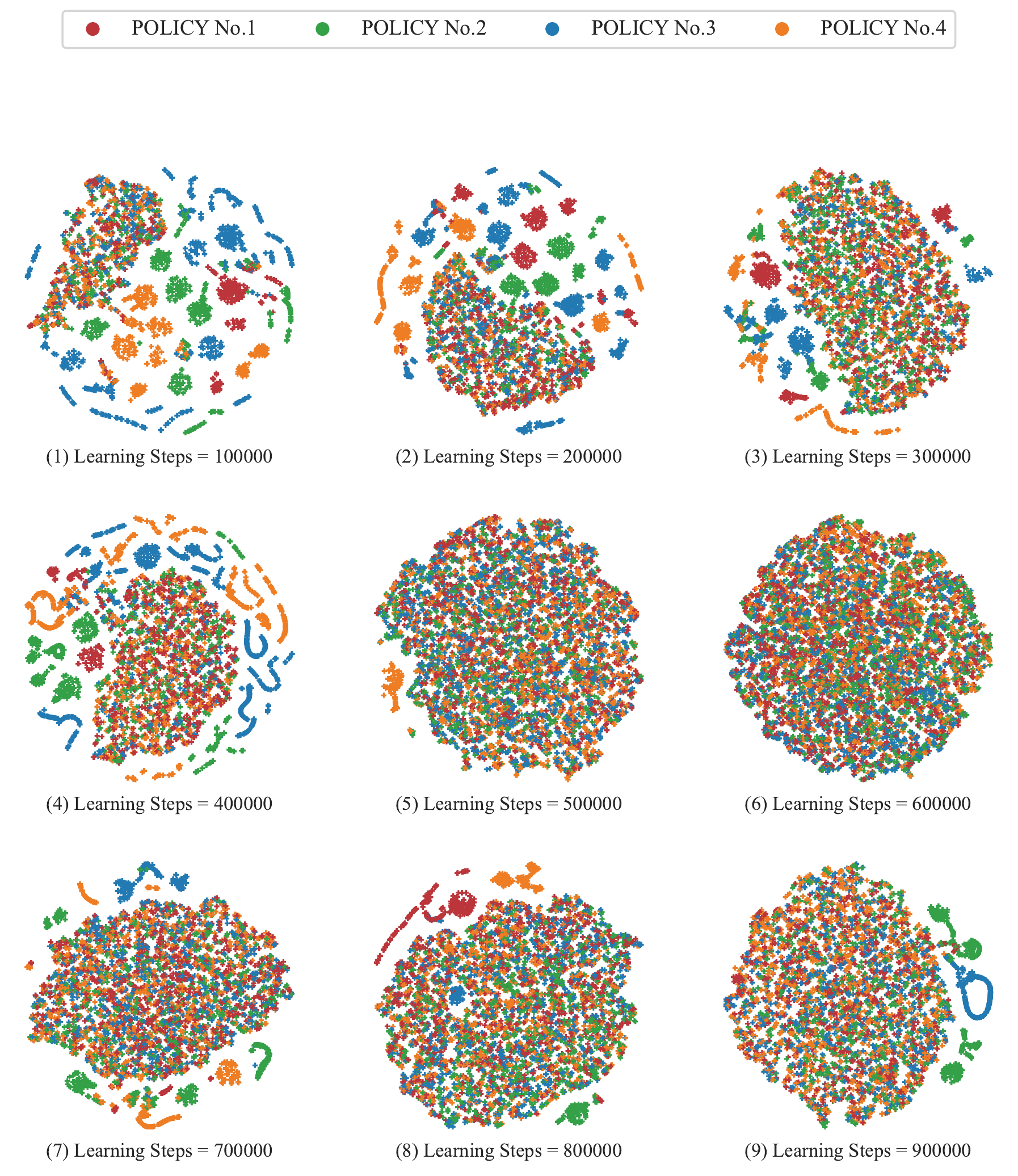}
    \caption{The states visited by each style. For better visualization, the states get dimension reduction by t-SNE. The points with different color represents the states visited by the policy with the style. The distance between points represents the difference between states.}
    \label{fig:stylel}
\end{figure*}

We provide supplementary results for Section \ref{sec:style}. Figure \ref{fig:stylel} shows the states visited by each style over 1M time steps with intervals of 100k. The results show that different styles get consistent but new styles emerges as well, which brings enduring exploration capabilities.

\end{document}